\documentclass{article}

\usepackage{spconf,amsmath,graphicx,hyperref}

\usepackage{cite}
\usepackage{amsmath,amsthm}
\usepackage{enumitem}
\usepackage{amssymb}
\usepackage{graphicx}
\usepackage{textcomp}
\usepackage{xcolor,verbatim}
\usepackage{bbm}
\usepackage{dsfont}
\usepackage{subcaption}
\usepackage{url}
\usepackage{hyperref}
\usepackage{cleveref}
\usepackage[font=small]{caption}
\usepackage{algorithm} 
\usepackage{algorithmic}  
\usepackage[ruled,vlined,algo2e]{algorithm2e}

\usepackage{booktabs,makecell,multirow,hhline,threeparttable} 
\usepackage{xcolor,colortbl,pgf}
\definecolor{Color1}{RGB}{240, 240, 240}
\usepackage{makecell}

\usepackage{dsfont}
\usepackage{amssymb} 
\usepackage{bm} 
\usepackage{xspace}

\newcommand{\reals}{\mathbb{R}}
\newcommand{\ff}{\mathcal{F}}
\newcommand{\x}{\mathbf{x}}
\newcommand{\y}{\mathbf{y}}
\newcommand{\s}{\mathbf{s}}
\newcommand{\loss}{\mathcal{L}}

\title{Learning Nonlinear Systems In-Context: From Synthetic Data to Real-World Motor Control}

\name{Tong Jian\thanks{\copyright\quad2026 IEEE. Personal use of this material is permitted.  Permission from IEEE must be obtained for all other uses, in any current or future media, including reprinting/republishing this material for advertising or promotional purposes, creating new collective works, for resale or redistribution to servers or lists, or reuse of any copyrighted component of this work in other works.}\thanks{To appear in Proceedings of the IEEE International Conference on Acous-tics, Speech, and Signal Processing (ICASSP), 2026} \qquad Tianyu Dai \qquad Tao Yu}
\address{\normalsize{Analog Devices, Inc., Boston, MA, USA} \\ 
\normalsize{\{Tong.Jian, Tianyu.Dai, Tao.Yu\}@analog.com}
}

\begin{document}
\maketitle

\begin{abstract}
LLMs have shown strong in-context learning (ICL) abilities, but have not yet been extended to signal processing systems. Inspired by their design, we have proposed for the first time ICL using transformer models applicable to motor feedforward control, a critical task where classical PI and physics-based methods struggle with nonlinearities and complex load conditions.
We propose a transformer-based model architecture that separates signal representation from system behavior, enabling both few-shot finetuning and one-shot ICL. Pretrained on a large corpus of synthetic linear and nonlinear systems, the model learns to generalize to unseen system dynamics of real-world motors only with a handful of examples. In experiments, our approach generalizes across multiple motor–load configurations, transforms untuned examples into accurate feedforward predictions, and outperforms PI controllers and physics-based feedforward baselines. These results demonstrate that ICL can bridge synthetic pretraining and real-world adaptability, opening new directions for data-efficient control of physical systems.
\end{abstract}

\begin{keywords}
in-context learning, feedforward motor control
\end{keywords}

\section{Introduction}\label{sec:introduction}
\begin{figure}[t]
    \centering
    \includegraphics[width=0.9\columnwidth]{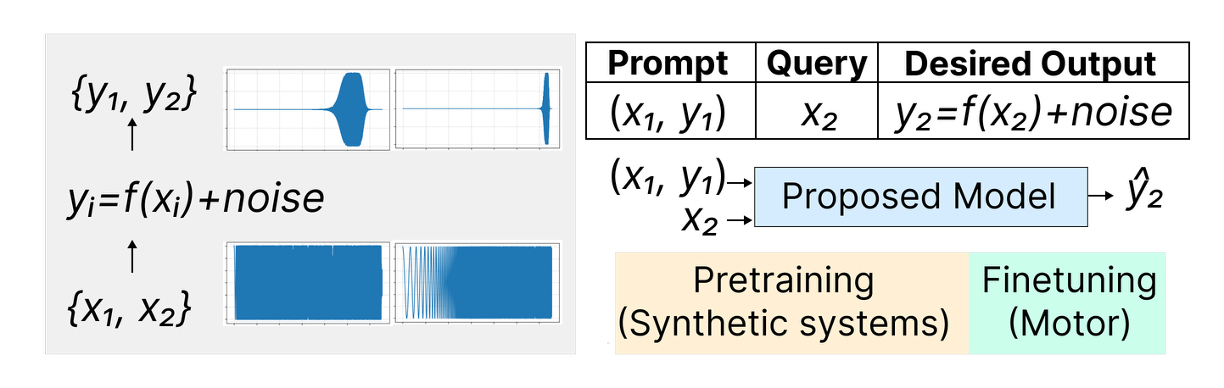}
    \vspace{-10pt}
    \caption{Proposed workflow. We consider systems $f$ that generate input–output pairs. The proposed model predicts a query output from one single pair in an in-context learning manner. Pretrained on a large corpus of synthetic linear and nonlinear systems and finetuned on a few motor examples, the model has the ability to generalize effectively to unseen motors and configurations.}
    \label{fig:intro}
\end{figure}

Modern transformers can adapt to new tasks by conditioning on example input–output pairs presented in the prompt. This ability, known as in-context learning (ICL), allows models to perform implicit adaptation without explicit parameter updates~\cite{dong2022survey, Guo2024, ICLR2024_45ed1a72, NEURIPS2024_75b0edb8, gu-etal-2023-pre}. Figure \ref{fig:intro} illustrates a common setup, where a model predicts a query output after seeing one single (input, output) pair as the prompt. Recent work has shown transformers can infer unseen linear functions from examples~\cite{garg2022simplefuntion, Li2023asalgorithms} and the promise on synthetic linear time-invariant (LTI) and Wiener–Hammerstein systems~\cite{piga2024adaptation, forgione2023systemmodel, piga2024syntheticSI, Colombo2025}. However, these efforts remain largely system-specific and simulation-only, leaving open the question: \textit{can ICL generalize to real-world physical systems?}

\begin{figure}[!t]
    \centering
    \includegraphics[width=1\columnwidth]{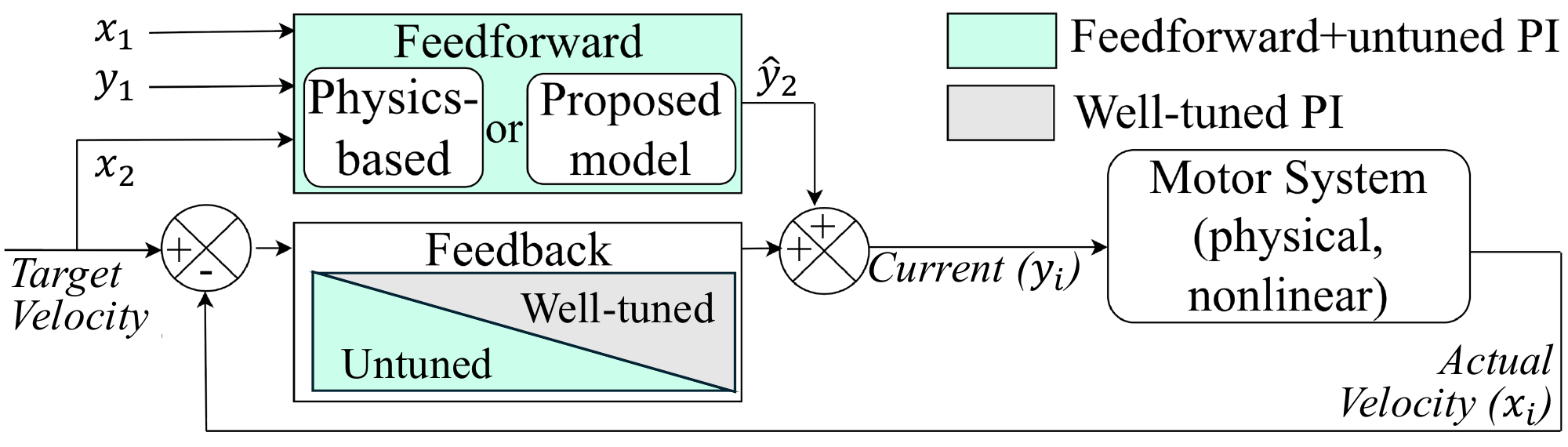}
    \vspace{-15pt}
    \caption{Block diagram of the motor control loop. A feedforward path complements the PI feedback controller to track the target velocity, reducing the tuning burden and enabling good performance on untuned motors. Our proposed model replaces the traditional physics-based method to predict effective feedforward signals. This removes the need for accurate motor system modeling and generalizes more easily to complex motor systems. }
    \vspace{-5pt}
    \label{fig:control_loop}
\end{figure}

Motor control is a compelling testbed for this question. As illustrated by the motor control loop diagram in Figure~\ref{fig:control_loop}, a proportional–integral (PI) feedback loop regulates the motor to track the desired motion, while a feedforward path provides a compensatory signal, computed in advance by predicting the required actuation to improve tracking performance. PI control, while simple and robust, requires manual tuning and struggles with nonlinearities. Physics-based feedforward can improve tracking but depends on highly accurate models, which are difficult to obtain for complex systems and often fail to capture unmodeled effects~\cite{johnson2002experimental}. Recent machine learning approaches have shown promise for enhancing motor feedforward control~\cite{bolderman2023data, liu2019industrial, kazemikia2024reinforcement, busetto2024meta}, yet they generally require large datasets or retraining for each new motor or load. A crucial but desired next step is to develop models that can infer system dynamics from only a few--or even a single--untuned measurement and directly generate the feedforward signal needed to achieve the desired motion.

In this work, we explore what ICL can capture about nonlinear physical systems by bridging synthetic simulations with real-world motor control. Figure \ref{fig:intro} illustrates our workflow: a transformer-based model is pretrained on a large corpus of synthetic linear and nonlinear systems and then finetuned for motor feedforward control using only a few input–output pairs. Figure \ref{fig:control_loop} highlights the proposed usage: after finetuning, the model takes a single untuned measurement pair as a prompt and predicts the feedforward torque current for a desired target velocity. This approach enables rapid deployment across diverse hardware platforms with minimal manual intervention. Our main contributions are:
\begin{enumerate}[nosep]
    \item We introduce, for the first time, ICL for motor feedforward control, tackling nonlinearities and complex loads where classical PI and physics-based methods struggle. 
    \item We propose a transformer-based model supporting few-shot finetuning and one-shot ICL, excelling in both system representation and signal generation.
    \item We experimentally show that pretraining on synthetic data enables efficient adaptation to real-world motors. The model generalizes across unseen motor-load and outperforms both PI and physics-based baselines.
\end{enumerate}

The rest of the paper is organized as follows. Section \ref{sec:methodology} introduces the proposed model architecture and training strategy. Section \ref{sec:synthetic-data} and \ref{sec:pretrain} describe the synthetic data generation and pretraining results. Section \ref{sec:motor} discusses the motor setup and feedforward motor control results.

\section{Methodology}\label{sec:methodology}
\begin{figure}[!t]
    \centering
    \includegraphics[width=0.95\columnwidth]{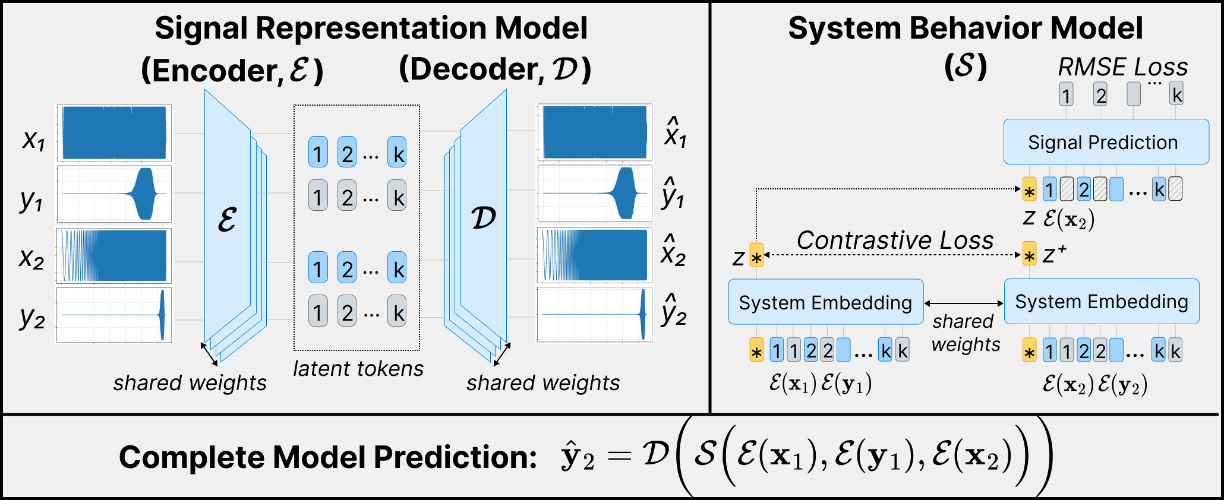}
    \vspace{-5pt}
    \caption{Proposed model architecture. The signal representation model is an encoder–decoder that encodes both $\x$ and $\y$ into token sequences. The system behavior model has two blocks: a system embedding block, which learns a system-level embedding $z$ that characterizes system behavior from pairs, and a signal prediction block, which uses $z$ as the prompt together with tokens of $\x_2$ to generate accurate $\hat{\y}_2$ predictions.}
    \label{fig:model}
    \vspace{-5pt}
\end{figure}

We proposed a generalized model with two-stage pretraining: signal representation and system behavior learning. The architecture (see Figure~\ref{fig:model}) is designed to (a) excel in system understanding and generation, (b) enable one-shot in-context learning, and (c) support real tasks in a data-efficient manner.

\subsection{Problem Formulation}
We consider a family of systems $\ff = \{f^i\}_{i=1}^N$, where each system maps finite-length input sequences to output sequences. For a system $f^i \in \ff$, let $(\x^i_j, \y^i_j)$ denote its $j$-th input-output pair, with $\x^i_j \in \reals^T, \y^i_j \in \reals^T$ and $\y^i_j = f^i(\x^i_j)$. 
Given any $f^i \in \ff$ and an observed pair $(\x_1^i, \y_1^i)$, the goal is to learn a model that generalizes across $\ff$ by capturing system behavior directly from input–output examples—without explicitly recovering the underlying functional form—and accurately predicting $\y_2$ for a new input $\x_2$.

\subsection{Stage 1: Signal Representation Learning}\label{sec:stage1}
The signal representation model is designed as an encoder- decoder that processes input $\x$ and output $\y$ signals individually. The encoder maps a raw signal into a sequence of latent tokens capturing its essential features, while the decoder reconstructs the signal from these tokens. We adopt Encodec \cite{encodec} as the backbone for its proven strong performance on audio signal representation and lightweight design. Unlike the original architecture, we omit the discriminator and retain only the encoder, quantizer, and decoder. Using 8 codebooks with embedding dimension 128 and an encoder downsampling ratio of 320, a signal $\s$ of length T can be represented as a latent tensor of shape $[\lceil T/320 \rceil, 128]$.

Training is performed using a reconstruction loss based on mean squared error (MSE): $ \loss_{signal} = \frac{1}{M} \sum_{i=1}^M | \s_i - \hat{\s_i} |^2$,
where $\s$ and $\hat{\s}$ denote the raw and reconstructed signals, respectively, and $M$ is the number of training signals. Overall, this design ensures that encoding remains length-adaptive, allowing variable-length signals to be represented by proportional token sequences without resampling. It also leverages self-supervised pretraining to improve generalization at the signal level, enabling the system behavior model to concentrate solely on learning inter-signal dynamics.

\subsection{Stage 2: System Behavior Learning}\label{sec:stage2}
Leveraging pretrained signal tokens, the system behavior model is trained to capture system characteristics from input–output pairs. The model consists of two coupled blocks. The system embedding block uses self-attention layers with relative positional encodings to process paired tokens and a learnable system token, producing an embedding $z^i$ that characterizes system behavior. A contrastive loss structures the embedding space, pulling embeddings of the same system closer while pushing apart those from different systems:

\begin{equation}\label{eq:system-contra}
    \loss_{contra} = -log \frac{exp(s(z^i, z^{i,+})/\tau)}{\sum_{j=1}^N exp(s(z^i, z^j)/\tau)} 
\end{equation}

where $\tau$ is the temperature parameter and $s(u,v) = \frac{u^Tv}{||u||||v||}$ denotes cosine similarity. When training in a batch, positives are the pairs from the same system, and negatives are taken from other pairs within the training batch.

The signal prediction block takes the system embedding $z^i$ learned from the first pair and stacks it with the query tokens $\x^i_2$ via self-attention layers and relative positional encodings to predict the corresponding output $\hat{\y}^i_2$. A reconstruction loss ensures fidelity to the true output $\y_2^i$, ensuring that the system embedding guides the prediction effectively.
\begin{equation}\label{eq:system-recon}
    \loss_{recon} = \frac{1}{N} \sum_{i=1}^M | \y^i_2 - \hat{\y^i_2} |^2
\end{equation}

The overall system behavior model is optimized under a joint objective that combines the contrastive embedding loss with the reconstruction loss.
Through this design, the embedding block produces a system-level prompt that supports one-shot in-context learning and generalization to unseen systems, while the prediction block ensures that the learned prompt can be effectively applied for accurate input–output inference.

\subsection{Downstream Task Adaptation}
The signal representation and system behavior models are pretrained sequentially on synthetic data, establishing generalizable foundations for both signals and systems. With this initialization, model adapts to motor tasks using only a few pairs by finetuning a single layer of the system behavior model. Once finetuned, it generalizes to unseen motors in a one-shot manner: from a single untuned pair as a prompt, the model infers the new system behavior that acts as a motor–setup indicator, enabling accurate predictions for new target velocity waveforms. This two-stage design enables data-efficient adaptation, supports in-context learning, and ensures robust generalization to previously unseen systems.

\section{Experiments}\label{sec:experiments}
In this section, we explore how the proposed generalized model extends to real-world motor control and how various pretraining strategies affect its performance. Our study is guided by two key questions: (1) Under what scenarios does the proposed model enhance motor control performance? (2) How do different choices of pretraining data influence downstream task outcomes?

\begin{figure}[!t]
    \centering
    \includegraphics[width=0.95\columnwidth]{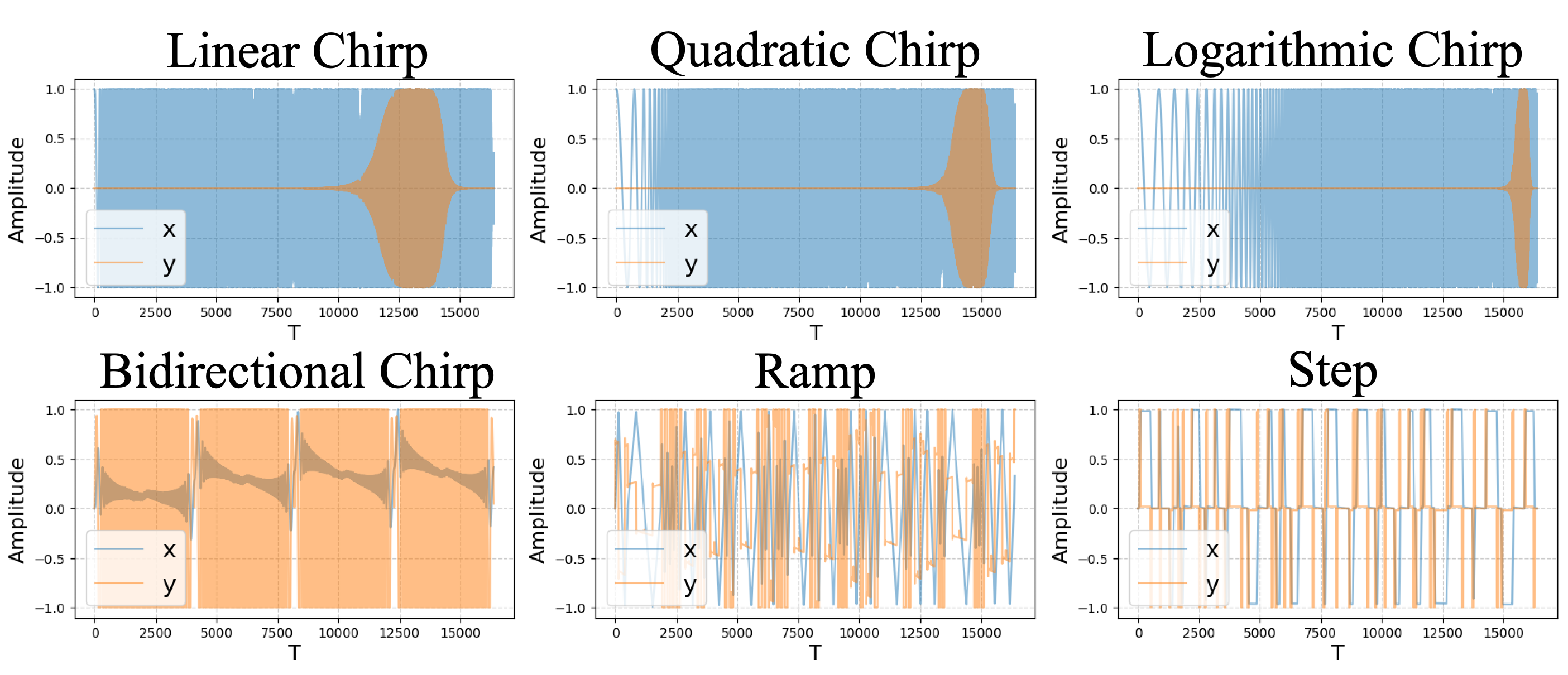}
    \vspace{-5pt}
    \caption{Synthetic dataset examples: input–output pairs with six input signal types. The first row shows pairs generated by a third-order LTI bandpass filter. The second row shows pairs generated by a third-order NTI system with saturation and static friction.}
    \vspace{-10pt}
    \label{fig:dataset}
\end{figure}

\begin{figure}[t]
    \centering
    \includegraphics[width=0.95\columnwidth]{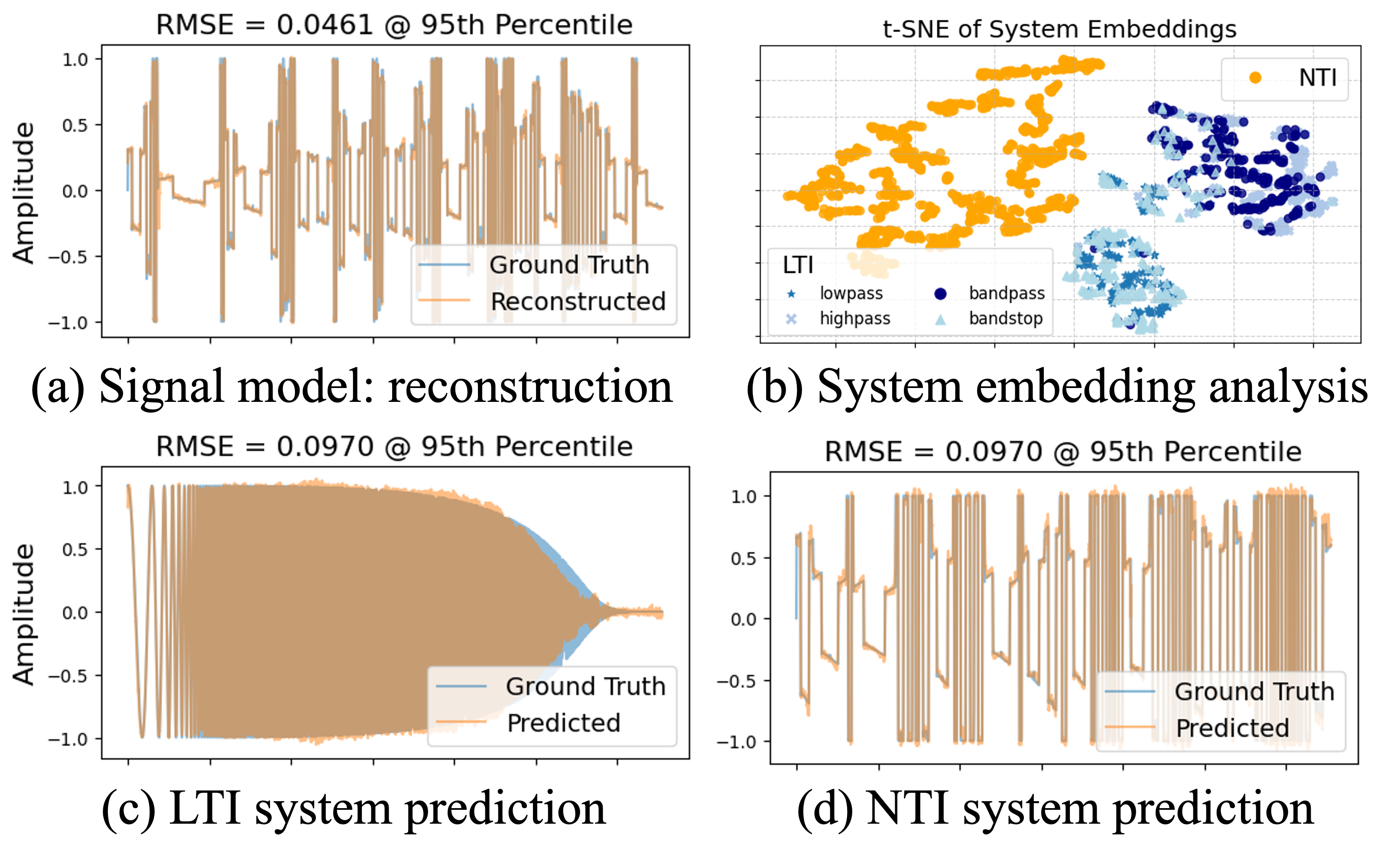}
    \vspace{-5pt}
    \caption{Pretraining performance: (a) signal representation model reconstruction at the 95th percentile RMSE, (b) system embeddings visualized with t-SNE, showing clear separation between LTI and NTI systems, and further distinction among filter types (lowpass/bandstop vs. highpass/bandpass), and (c)-(d) system prediction examples for both LTI and NTI system at the 95th percentile RMSE.}
    \vspace{-10pt}
    \label{fig:pretrain}
\end{figure}

\subsection{Synthetic Data Generation}\label{sec:synthetic-data}
We constructed a diverse dataset of 60,000 input–output signal pairs from 20,000 synthetic systems, with example pairs shown in Fig.~\ref{fig:dataset}. All signals were normalized in frequency and set to a length of 16,384. The dataset includes both linear time-invariant (LTI) and nonlinear time-invariant (NTI) systems. LTI systems were generated using the SciPy filter design toolbox, spanning orders 1–4 and filter types including lowpass, highpass, bandpass, and bandstop, with randomized hyperparameters for generalizable dynamics. 

NTI systems were generated using a custom simulator that builds on random third- or fourth-order LTI models and augments them with nonlinearities. The base LTI models were constrained for stability through positive gains, bounded damping, and limited bandwidth. Deadzones with small random offsets and symmetric saturation with random limits were incorporated to capture practical non-ideal behaviors. These randomized elements produce diverse and realistic input–output trajectories that reflect the nonlinear and constrained dynamics of real electromechanical systems. 

The dataset was split into 18,000 systems for training and 2,000 for testing, ensuring evaluation only on unseen systems. We emphasize that the system metadata and governing equations were never used; the model learned system behavior solely from raw input–output pairs.

\subsection{Pretraining Results}\label{sec:pretrain}
We first trained the signal representation model using both input and output signals from the synthetic training set. Model quality is evaluated by reconstructing test signals, achieving a mean RMSE of 0.0201. This confirms that the model learns meaningful tokens. Qualitative results are shown in Fig.~\ref{fig:pretrain}(a).

With the signal model fixed, we then pretrained the system behavior model and evaluated its effectiveness. T-SNE projections of system embeddings (Fig.\ref{fig:pretrain}(b)) show that same types of systems cluster together, indicating the model captures key system characteristics. One-shot prediction on unseen systems (Figs.\ref{fig:pretrain}(c)–(d)) achieved a mean RMSE of 0.0469, demonstrating that the model can infer system behavior from a single input–output pair.

\begin{table}[t!]
\caption{Comparison with baselines. For ramp waveform as the target velocity, the predicted torque current is applied to the motor, and the resulting velocity is measured and compared with the target. The proposed model achieves the lowest RMSE.}
\vspace{-15pt}
\label{table:motor}
\begin{center}
\scalebox{0.7}{
\begin{tabular}{l||cc|c|c}
\toprule 
 \multirow{2}{*}{\textbf{Method}}  & \multicolumn{2}{c|}{\textbf{Unseen load}} & \multirow{2}{*}{\textbf{Unseen Motor}} & \multirow{2}{*}{\textbf{Two-inertia System}}  \\
 & BLDC & Stepper && \\
\midrule
\midrule
   Well-tuned PI & 1.36 & 1.74  & 1.91 & 3.64 \\
   Physics-based FF & \textbf{0.70} & 0.76 & 0.46 & 1.70 \\
\midrule
   Ours & \textbf{0.70} & \textbf{0.48} & \textbf{0.45} & \textbf{1.13} \\
\midrule
\end{tabular}
}
\end{center}
\vspace{-15pt}
\end{table} 

\begin{figure}[t]
    \centering
    \includegraphics[width=1\columnwidth]{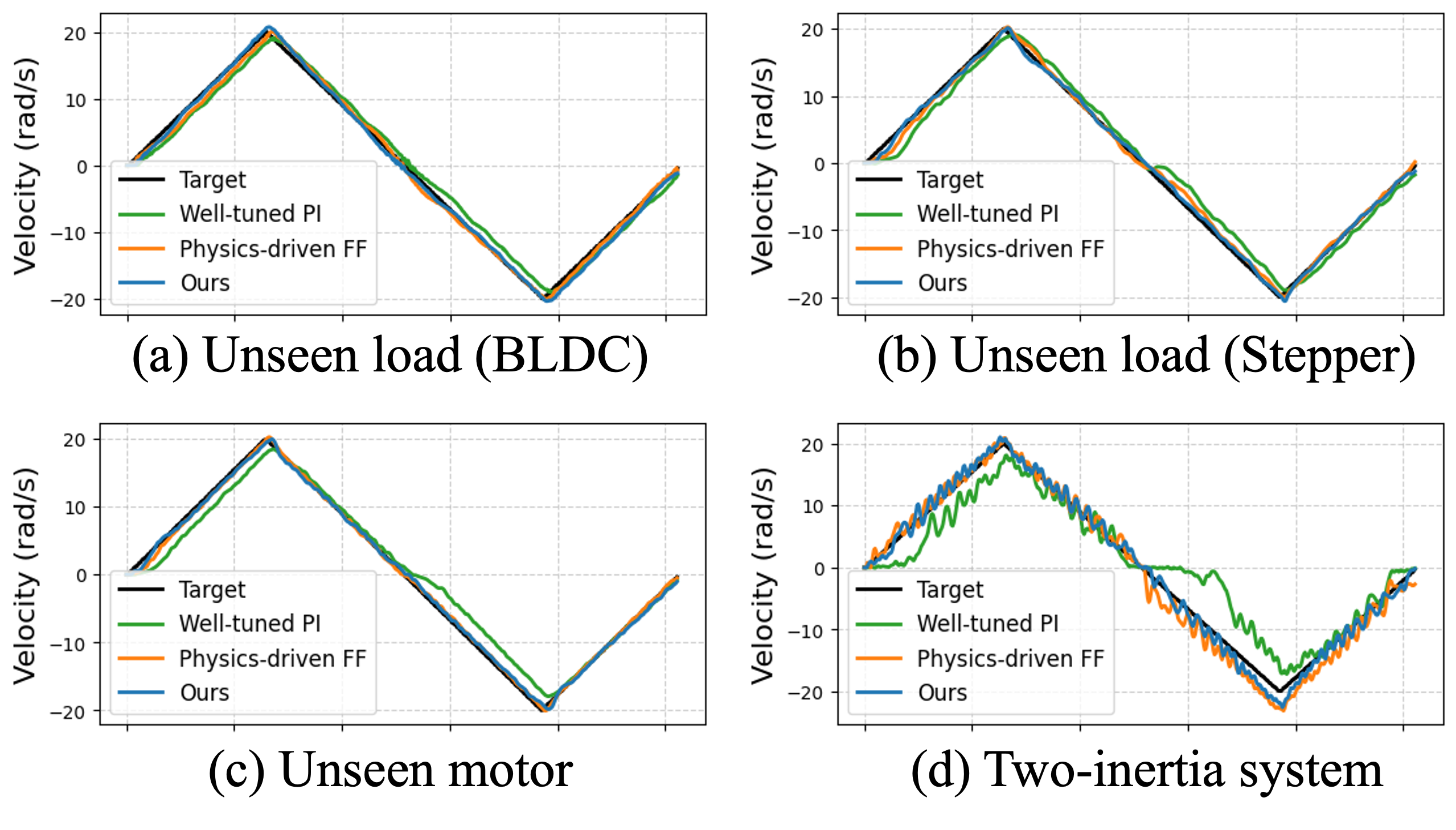}
    \vspace{-20pt}
    \caption{Qualitative comparison with baselines. }
    \label{fig:motor}
    \vspace{-10pt}
\end{figure}

\subsection{Real-world Motor Application}\label{sec:motor}
We evaluated our proposed approach on real-world feedforward control tasks using an Analog Devices Trinamic motor controller TMC9660 to drive both stepper and brushless DC (BLDC) motors under varied loads. Motor velocity $\omega$ and torque current $i_q$ were measured, with the goal of having a model that predicts $i_q$ required to achieve a given target $\omega$. 

For finetuning, we collected two types of $(\omega, i_q)$ pairs: untuned pairs from PI control and well-tuned pairs from PI control combined with physics-based feedforward (see Figure \ref{fig:control_loop}). The untuned pairs serve as the prompt, reflecting realistic usage where PI controllers are not pre-tuned and enabling the model to capture the system’s real dynamics. The well-tuned pairs serve as the query, guiding the model to generate the feedforward $i_q$ for accurate control. 
The signal encoder was kept frozen, with only one single layer of the system behavior model lightly finetuned for 20 epochs.

We evaluated several waveform types for the $\omega$ in the prompt, and found that chirp signals guided the most accurate predictions on unseen velocity waveform due to their wide frequency coverage (0.001–10 Hz). Therefore, all reported results use chirp prompts with ramp or step queries. We compare against two baselines: a well-tuned PI controller and a Physics-based modeling feedforward controller.


\textbf{Case 1 — Unseen load on known motors:}
We finetuned on one stepper and one BLDC motor on five single-inertia loads ranging from $0$ to $5\times10^{-4}~\mathrm{kg\cdot m^2}$, then tested on an unseen load of $1\times10^{-4}~\mathrm{kg\cdot m^2}$ within this range. Both quantitative (Table~\ref{table:motor}) and qualitative results (Figure~\ref{fig:motor}) suggest our method outperforms the PI and the physics-based feedforward control.

\textbf{Case 2 — Unseen motor:}
We took the model finetuned for Case 1 and directly tested on a new unseen stepper motor with entirely different parameters. Despite having seen data from only two motors during finetuning, the model achieved accurate predictions on the unseen motor, demonstrating strong cross-motor generalization. 

\textbf{Case 3 — Two-inertia system:}
We further evaluated the model on a two-inertia system, where a motor drives a linear slide under different loads. Remarkably, finetuning on only two load conditions (zero and heavy) enabled accurate prediction for an unseen light load. While this setup is challenging for physics modeling, our approach outperforms both the PI and the physics-based feedforward control, demonstrating strong generalization to complex real-world dynamics. 

\vspace{-5pt}
\subsection{Ablation Studies}
We use Case 3 to study the impact of 2-stage pretraining and pretraining data on downstream performance. 
Empirically, 2-stage training consistently outperforms joint end-to-end training, achieving an RMSE of 1.13 vs. 1.20 for joint training on ramp queries. We further compare the model pretrained on both LTI and NTI pairs with the model pretrained on LTI pairs only. For ramp queries, the LTI-only model achieves an RMSE of 1.17 compared to 1.13 for the LTI+NTI model; for step queries, it achieves 2.16 versus 1.79. While adding NTI generally boosts performance, the LTI-only model remains competitive and outperforms both baselines, suggesting that synthetic pretraining data need not exactly match the target system to be effective.



\vspace{-5pt}
\section{Conclusions}\label{sec:conclusion}
\vspace{-5pt}
We introduced a novel framework for in-context learning on time-invariant nonlinear systems from synthetic data with demonstration of its generalizability adapted to real-world motor feedforward control problem using very few data samples.
The model generalizes across unseen motor–load configurations and surpasses commonly used PI and physics-based method. These findings demonstrate that synthetic pretraining, even when not perfectly matched to the target system, can enable efficient real-world adaptation—highlighting in-context learning as a powerful paradigm for data-efficient control of physical systems.
\vspace{-5pt}
\section{Acknowledgments}\label{sec:acknowledgments}
\vspace{-5pt}
We thank Colm Prendergast and Audren Cloitre for their insightful discussions that helped improve this work.
\bibliographystyle{IEEEbib}
\bibliography{ref}
\end{document}